\documentclass[runningheads]{llncs}

\usepackage{eccv}

\usepackage{eccvabbrv}

\usepackage{graphicx}
\usepackage{booktabs}
\usepackage{float}

\usepackage[accsupp]{axessibility}  

\usepackage{hyperref}

\usepackage{orcidlink}

\begin{document}

\title{VQA-Diff: Exploiting VQA and Diffusion for
Zero-Shot Image-to-3D Vehicle Asset Generation in Autonomous Driving}

\titlerunning{VQA-Diff}

\author{Yibo Liu\inst{1,2,}\thanks{ Equal contribution. Work done during an internship with Huawei Noah's Ark Lab.}\orcidlink{0000-0003-1143-3242}, 
Zheyuan Yang\inst{1,3,*}, 
Guile Wu\inst{1},
Yuan Ren\inst{1}\orcidlink{0000-0002-4901-3596},
Kejian Lin\inst{1},
Bingbing Liu\inst{1}\orcidlink{0000-0002-5272-3425},
Yang Liu\inst{1}, \and
Jinjun Shan\inst{2}\orcidlink{0000-0002-4911-6739}
}

\authorrunning{Y. Liu, Z. Yang et al.}

\institute{Huawei Noah’s Ark Lab, Toronto ON L3R 5A4, Canada \and
York University, Toronto ON M3J 1P3, Canada \and
University of Toronto, Toronto ON M5S 1A1, Canada\\
\email{ buaayorklau@gmail.com, andrewzheyuan.yang@mail.utoronto.ca, \{guile.wu, yuan.ren3, liu.bingbing, yang.liu9\}@huawei.com, jjshan@yorku.ca}}


\maketitle

\begin{figure}[H]
\hsize=\textwidth 
\centering
\includegraphics[width=12cm]{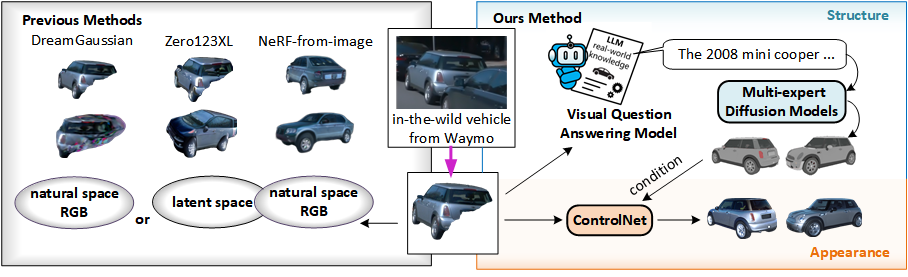}
\caption{Previous methods learn to generate novel views using image RGB information in a natural space or a latent space,
resulting in poor zero-shot prediction capability to handle in-the-wild vehicle observations with occlusion or tricky viewing angles. Our method, VQA-Diff, tackles this problem by exploiting
the robust zero-shot prediction ability of the Visual Question Answering (VQA) model and
the rich structure and appearance generation ability of Diffusion Models.
This helps to create consistent and photorealistic multi-view renderings of any unseen vehicle in the wild.
}
\label{mov}
\end{figure}


\begin{abstract}

Generating 3D vehicle assets from in-the-wild observations is crucial to autonomous driving.
Existing image-to-3D methods cannot well address this problem because they learn generation merely from image RGB information
without a deeper understanding of in-the-wild vehicles (such as car models, manufacturers, \etc). This leads to their poor zero-shot prediction capability to handle real-world observations with occlusion or tricky viewing angles.
To solve this problem, in this work, we propose VQA-Diff, a novel framework that leverages in-the-wild vehicle images to create photorealistic 3D vehicle assets for autonomous driving.
VQA-Diff exploits the real-world knowledge inherited from the Large Language Model in the Visual Question Answering (VQA) model
for robust zero-shot prediction and the rich image prior knowledge in the Diffusion model for structure and appearance generation.
In particular, we utilize a multi-expert Diffusion Models strategy to generate the structure information and employ a subject-driven structure-controlled generation mechanism to model appearance information.
As a result, without the necessity to learn from a large-scale image-to-3D vehicle dataset collected from the real world, VQA-Diff still has a robust zero-shot image-to-novel-view generation ability.
We conduct experiments on various datasets, including Pascal 3D+, Waymo, and Objaverse, to demonstrate that VQA-Diff outperforms existing state-of-the-art methods both qualitatively and quantitatively.

\end{abstract}

\section{Introduction}
\label{sec:intro}
Photorealistic 3D vehicle asset generation from in-the-wild images is important to autonomous driving as it benefits many downstream tasks, such as training data augmentation and developing sim2real technology \cite{sim1,sim2}.
It aims to create novel renderings of the vehicle given a single RGB image captured in the wild.
Although some image-to-3D methods \cite{nfi,zero,dg} may be used for photorealistic 3D vehicle asset generation,
these methods only learn to generate novel views from RGB information in the natural space and latent space without deeply understanding the characteristics of real-world vehicles.
As a result, they can merely learn from in-the-wild observations of vehicles collected in datasets \cite{p3d,obj}
that cannot cover all car models, manufacturers, occlusion and viewing angles observed in the real world.
As illustrated in Fig. \ref{mov}, when a given image does not fall into the distribution of the training data (\eg an image with occlusion or a tricky viewing angle), these methods cannot achieve effective zero-shot prediction with correct geometry and appearance to unseen observations.

Recently, the Visual Question Answering (VQA) model \cite{blip2} has shown impressive zero-shot prediction ability thanks to the broad real-world knowledge inherited from the Large Language Model (LLM) \cite{t5} and the rich image prior gained from the extensive visual training data \cite{clip,evaclip}.
In the application of autonomous driving, by setting questions to ask the VQA model about the car model, manufacturer, production year, and major features, we can obtain a detailed description of the vehicle even from an occluded in-the-wild observation as shown in Fig. \ref{mov}.
However, this zero-shot prediction ability only applies to image-to-text transformation and is thus not directly suitable for image-to-novel-views conversion.
On the other hand, Diffusion Models (DMs) \cite{Stable,blipd} are powerful generative models designed
for various text-to-image and image-to-image tasks.
A pretrained DM has abundant image prior and real-world knowledge to generate photorealistic images based on a prompt or image.
Despite the high fidelity of generated images, most DMs cannot control object poses for 3D asset generation.
\cite{zero} introduces the pose embedding into the original Stable Diffusion \cite{Stable} and employs it to rig the object pose in the output. However, as shown in the results of Zero123XL \cite{zero,obj} in Fig. \ref{mov}, relying merely on DMs cannot address the challenge of rendering novel views of in-the-wild vehicles because of out-of-distribution observations.

In this work, we develop a novel generative model, named VQA-Diff, which inherits the merits from both VQA and DMs, to tackle the problem of 3D vehicle asset generation from in-the-wild observations.
As illustrated in Fig. \ref{mov}, instead of directly learning image-to-3D or image-to-multi-view mappings,
VQA-Diff utilizes text with encoded broad real-world knowledge inherited from LLM as an intermediary to bridge the VQA model with DMs for 3D asset generation.
Considering that the zero-shot prediction of VQA only applies to image-to-text, we design multi-expert DMs to convert text into multi-view structures.
This multi-expert DMs design facilitates learning better image quality and vehicle structure compared to using a single DM.
Then, we employ the structures given by multi-expert DMs as the controlling condition and the raw image as the driving subject to generate the photorealistic appearance for multi-views with an edge-to-image ControlNet \cite{blipd,control}.
In this way, VQA-Diff does not have to learn the generation from a large-scale image-to-3D vehicle dataset collected from the real world
but maintains robust zero-shot image-to-novel-view generation ability.

The \textbf{contributions} of this work are threefold:
\begin{itemize}
\item We introduce a novel generative model, dubbed VQA-Diff, for creating photorealistic novel renderings from one single in-the-wild vehicle image. VQA-Diff utilizes the robust zero-shot prediction ability in the VQA model
and the rich structure and appearance generation ability in DMs for 3D asset generation.
\item We design a multi-expert DMs strategy to learn vehicle structures, which generates multi-views with better image quality and consistency compared to using a single DM.
\item We conduct both qualitative and quantitative experiments on three datasets, including Pascal 3D+ \cite{p3d}, Waymo \cite{waymo}, and Objaverse \cite{obj}, to demonstrate the superiority of our method over the existing state-of-the-art methods.
\end{itemize} \par

\par

\begin{figure}[tb]
  \centering
  \includegraphics[height=5.3cm]{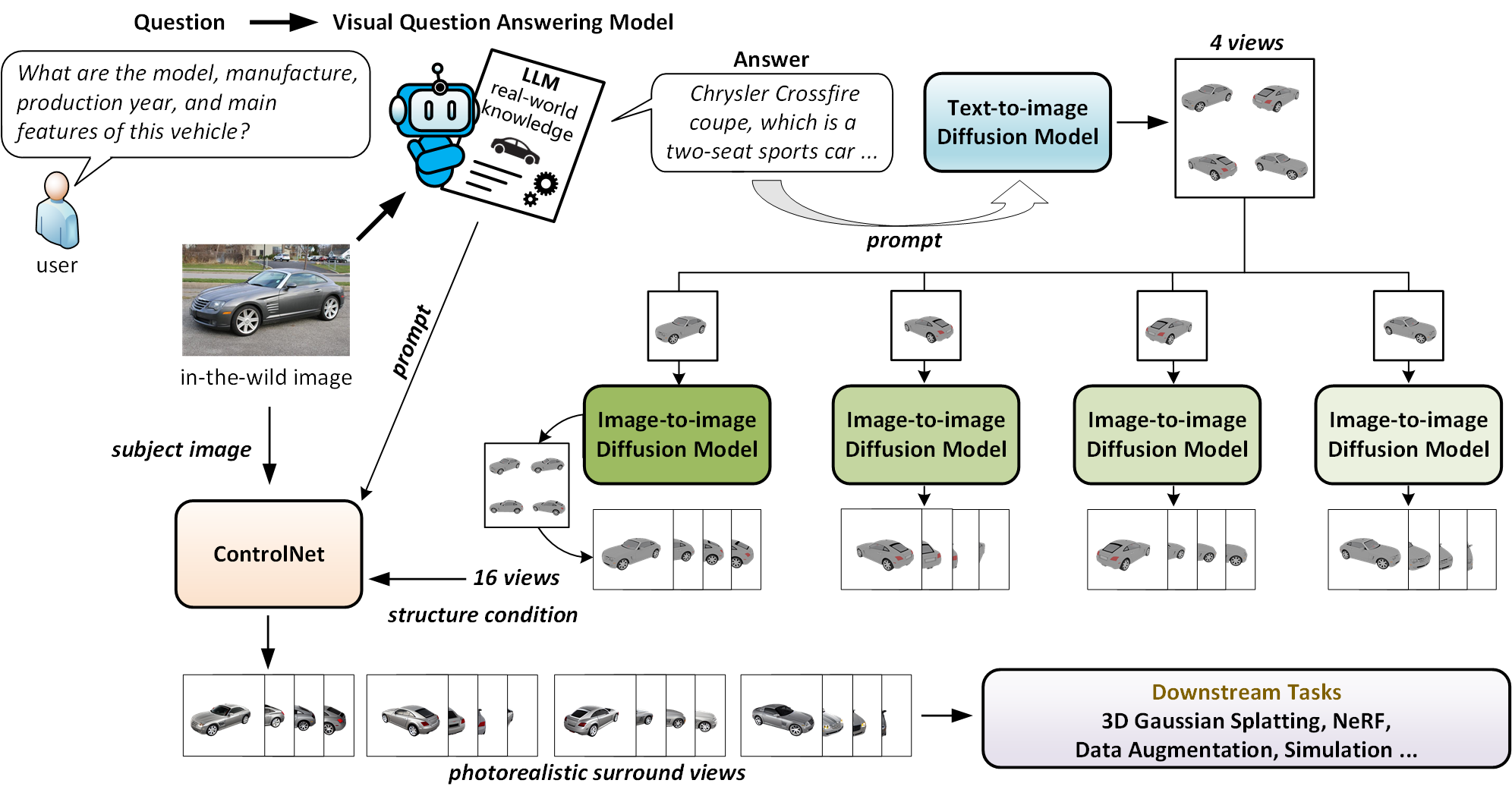}
  \caption{The framework of the proposed VQA-Diff. The VQA model first generates a prompt containing detailed key information regarding the model, manufacturer, production year, and main features of the vehicle. Then, multi-expert DMs adopt the prompt to create multi-view structures of the vehicle. Finally, the subject-driven structure-controlled generation with ControlNet renders the multi-view structures into photorealistic novel views with controllable poses. The photorealistic novel views can be utilized in various downstream tasks, including the creation of 3D assets with the GS/NeRF representation and training data augmentation. It can also be applied in a simulation environment for autonomous driving.
  }
  \label{framework}
\end{figure}

\section{Related Work} \label{related}
\noindent\textbf{Novel views synthesis from multi-view images.} Gaussian Splatting (GS) \cite{gs} and Neural Radiance Fields (NeRFs) \cite{nerf} are currently the most popular and widely used solutions for 3D reconstruction from multi-view images. Despite their different rendering strategies, their standard use-case is encoding/learning representation of a scene given multi-view images with associated camera poses and then rendering novel views.
%
%
Many follow-up works on GS and NeRFs are dedicated to improving rendering quality \cite{mip1,mip2,deblur}, reducing training time \cite{instant,compressed3d}, and extending to dynamic scenes \cite{4d1,4d2,dn,dn2,dn3}. Moreover, there has been some work \cite{sparseneus,reconfusion,sparsefusion,pixelnerf} focusing on reducing the number of required images.

\noindent\textbf{Novel views synthesis from a single image.} 
A common solution, as adopted in \cite{get3d,nfi,lolnerf,codenerf,gstriplane}, for synthesizing novel views from a single image is to learn the generalizable backbones of NeRFs and GS by encoding each object/scene with a latent code. Among them, NeRF-from-Image (NFI) \cite{nfi} develops a framework to learn the generation of shape, pose, and appearance of vehicles on Pascal 3D+ \cite{p3d}, which is a dataset containing posed single views of vehicles collected from the real world. Unlike these methods leveraging 3D representation, Zero123 \cite{zero} introduces pose embeddings into the original 2D Stable Diffusion (SD) \cite{Stable}, aiming to generate novel views directly from an image-to-image perspective. Some work \cite{make,dg,one2345, wonder3d} utilizes the diffusion prior in assisting 3D content generation through Score Distillation Sampling (SDS). For example, DreamGaussian (DG) \cite{dg} is the first work introducing diffusion prior from Zero123 \cite{zero} into 3D content generation with the GS representation. The aforementioned methods learn the image-to-novel-view mapping merely using the image RGB information in the natural space and latent space. They fail to deeply understand the characteristics of vehicles with real-world knowledge. Thus, their performance is limited to training samples included in datasets \cite{obj,p3d}.
Unfortunately, existing datasets \cite{obj,p3d} do not contain sufficient car models, various viewing angles, and complex occlusion cases. The incomplete representation of vehicles in existing datasets will result in a failure of the previous methods in developing their zero-shot prediction capability for novel view rendering. We intend to learn from a deeper understanding of the vehicles to achieve better novel views rendering performance. Particularly, we propose to transfer the robust zero-shot prediction ability of the VQA model \cite{blip2} into the generation of novel views in this work.


\section{Methodology}
The overview framework of VQA-Diff is depicted in Fig. \ref{framework}. There are three components: the VQA model, multi-expert DMs, and the ControlNet. We render novel views from a single image with three steps, the VQA processing, the structure generation, and the appearance generation. The VQA processing introduced in Sec. \ref{vqa} transforms the in-the-wild observation into a detailed prompt containing key information. Then, the multi-expert DMs proposed in Sec. \ref{multi} generate the consistent multi-view structures of the vehicle. Finally, as introduced in Sec. \ref{appear}, an edge-to-image ControlNet is utilized to render the multi-view structures into photorealistic novel views.  
\par

\begin{figure}[tb]
  \centering
  \includegraphics[height=3.6cm]{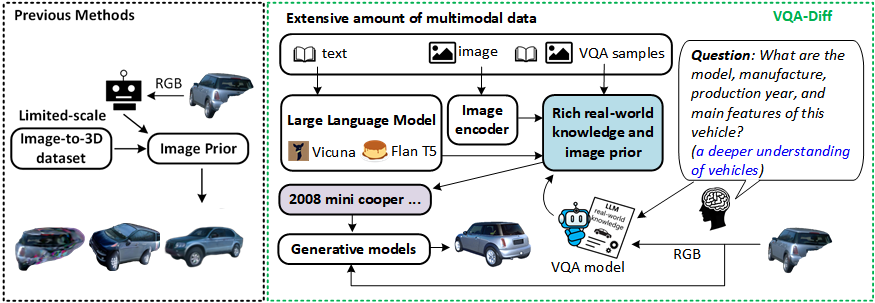}
  \caption{A comparison of the processes for dealing with the image-to-novel-view problem of previous methods and the proposed VQA-Diff. 
  }
  \label{com}
\end{figure}

\subsection{VQA Processing} \label{vqa}
\noindent\textbf{Motivation for VQA processing.} Considering the complex structures and appearances of vehicle observations in autonomous driving, the model must have a robust zero-shot prediction ability to render novel views. As aforementioned, it is tricky to develop the zero-shot prediction ability through learning an image-to-novel-view mapping due to the lack of an ideal large-scale dataset. Thus, instead of only focusing on the image modal and developing the zero-shot prediction ability from scratch, we opt to introduce the modal of text and integrate the zero-shot prediction ability of the VQA model into this problem. In particular, the text in the VQA \cite{blip2,vqa2} model is encoded with rich real-world knowledge inherited from the LLMs \cite{opt,t5}, which are trained with a tremendous amount of text samples. Furthermore, the image encoder \cite{clip,evaclip} of the VQA model also gained rich image prior from extensive image samples. The VQA model utilizes the strengths of LLMs and the image encoder through the design of bridging modules. These modules are trained using a comprehensive set of VQA samples (\eg BLIP-2 \cite{blip2} is trained with 254 million VQA samples). Thus, VQA models can robustly extract information from an image with a deeper understanding than the models that merely gain image prior from scale-limited image-to-3D datasets (\eg Objaverse-XL \cite{obj} contain around 10 million objects and Pascal 3D+ \cite{p3d} only covers 8500 instances). Leveraging the better/deeper image-understanding ability of the VQA models, we can extract useful and detailed vehicle information to boost the novel view generation of vehicles. A comparison of the processes for handling the image-to-novel-view problem of previous methods \cite{nfi,zero,dg} and the proposed VQA-Diff is presented in Fig. \ref{com}. 
\par
\begin{figure}[tb]
  \centering
  \includegraphics[height=3.0cm]{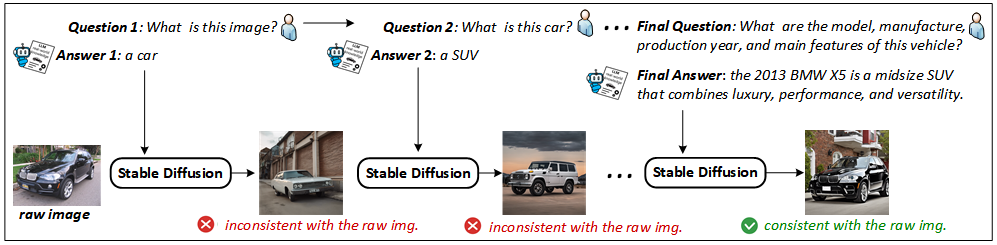}
  \caption{A illustration of the question design. We tune the question based on the feedback from Stable Diffusion \cite{Stable}.
  }
  \label{design}
\end{figure} \par
\noindent\textbf{Question Design.} Although the VQA model \cite{blip2} inherits real-world knowledge from LLM \cite{t5}, designing the question is important as the answer is the prompt for the following generative models and the prompt is crucial for text-guided generation \cite{prompt}. Inspired by \cite{prompt}, we design the question based on the feedback of an SD model \cite{Stable}. 
As shown in Fig. \ref{design}, we first set a simple question "What is this image?" for the VQA model. The given answer is "a car". If we input this rough description into a pretrained SD, the output image is an old-fashioned sedan, which is inconsistent with the raw image. Thus, in the second step, we make the question more specific and ask the VQA model "What car is it?". This time the generated answer is "an SUV". Again we input the answer into the SD and obtain an image of a Jeep-like full-size SUV, which is still inconsistent with the raw vehicle. We keep adjusting the question to make the output image of the SD more and more consistent with the raw image. Finally, the designed question in this work is "What are the model, manufacture, production year, and main features of this vehicle?" With this question, the output of SD is exactly the same vehicle as in the one in the raw image.
More discussions on the question design can be found in the supplementary material.

\subsection{Multi-expert DMs for Structures Generation} \label{multi}
The geometry of a vehicle is determined if the key information including model, manufacturer, production year, and main features are given. Thus, the VQA model deals with the disocclusion of geometry by providing a detailed and accurate description. Since structure and appearance generation are separately handled for novel view rendering, VQA-Diff does not have to learn the generation of geometry and texture simultaneously as in previous methods \cite{nfi,zero,dg}. Instead, our model only learns to transform the prompt into structures at this stage. 
Inspired by the good generalizability of the previous work on shape completion of vehicles \cite{mending,mv,pointr} trained on ShapeNetV2 \cite{shapenet}, our model learns vehicle structures from the ShapeNetV2 dataset to transfer the zero-shot prediction of the VQA model to structures.

\par
\begin{figure}[tb]
  \centering
  \begin{subfigure}{0.45\linewidth}
  
    \includegraphics[height=3.5cm]{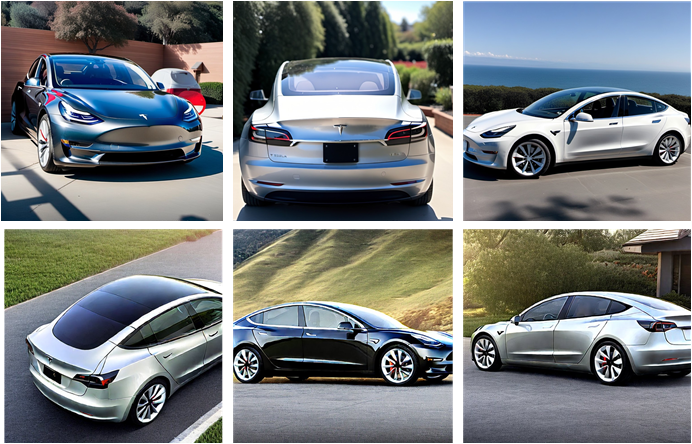}
    \caption{Images of Tesla Model 3 generated by a pretrained Stable Diffusion \cite{Stable}.}
   
    \label{sdt}
  \end{subfigure}
  \hfill
  \begin{subfigure}{0.5\linewidth}
    \includegraphics[height=3.5cm]{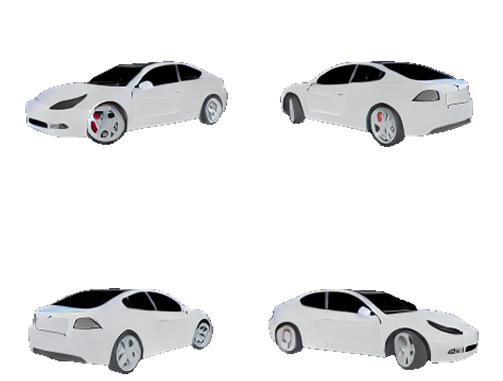}
    \caption{The multi-view structures of the Tesla Model 3 generated by our method. Note that our model does not learn the geometry of this car from ShapeNetV2 \cite{shapenet}.}
    \label{ourt}
  \end{subfigure}
  \caption{An illustration of transferring real-world knowledge and image prior of a pretrained DM \cite{Stable} into multi-view structure generation.}
  \label{dmt}
\end{figure}

 
\noindent\textbf{Motivation for adopting DM.} 
ShapeNetV2 \cite{shapenet} does not include the car models developed in recent years, such as the Tesla Model 3. To increase the variety of our model in asset creation, we utilize a pretrained SD model \cite{Stable} that has sufficient prior knowledge of vehicle structures. For example, when provided with a text prompt about the Tesla Model 3, the pretrained SD model can generate diverse and accurate structures of the vehicle, as shown in Fig. \ref{sdt}. However, the lack of control over vehicle poses in the output of SD hinders the utilization of the images as 3D vehicle assets. Thus, we fine-tune a pretrained SD model on the ShapeNetV2 dataset to control vehicle poses in the output while maintaining the model's capability of generating structures for various vehicles.
\par
\begin{figure}[tb]
  \centering
  \includegraphics[width=11.0cm]{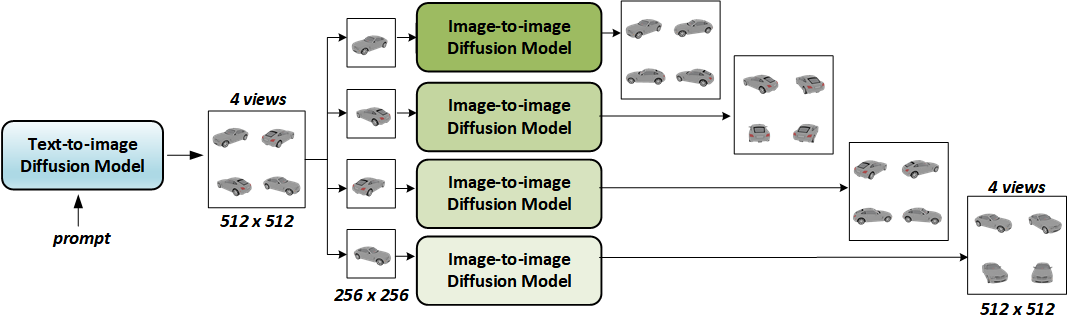}
  \caption{An illustration of the proposed multi-expert DMs.
  }
  \label{medm}
\end{figure}
\noindent\textbf{Design of Multi-expert DMs.} 
Fig. \ref{medm} shows an architecture of the multi-expert DMs design for vehicle structure learning. In particular, to transfer the zero-shot prediction of the VQA model, 
we first train a text-to-image DM that can generate a 512 $\times$ 512 image consisting of four 256 $\times$ 256 sub-images based on a text prompt, each of which contains a rendering of the vehicle from a different fixed camera pose. 
Due to multiple anchor views of these sub-images, the VQA model can learn structures from a wide variety of perspectives for a single car.
For each of the sub-images, we employ an image-to-image DM to generate a 512 $\times$ 512 image consisting of four 256 $\times$ 256 sub-images, resulting in a total of 16 surrounding views of the vehicle, with fixed and controllable camera poses equally spaced around the object. The design enables the model to capture the correlated local structures among the anchor views. 
Fig. \ref{ourt} shows the multi-view structure of the Tesla Model 3 generated by our method. As can be seen, despite the absence of this vehicle in ShapeNetV2, the VQA model successfully generates consistent images for the Tesla Model 3 in multiple views.

\begin{figure}[tb]
  \centering
  \begin{subfigure}{0.45\linewidth}
  
    \includegraphics[height=4.0cm]{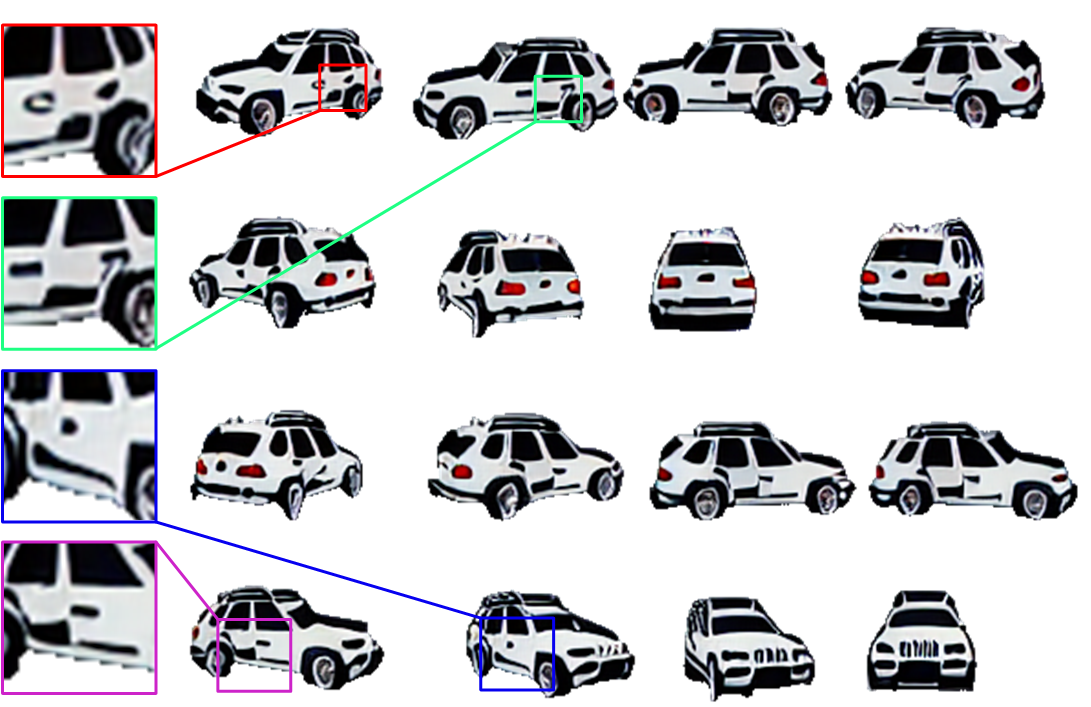}
    \caption{Multi-view structures generated by one single DM. }
   
    \label{161}
  \end{subfigure}
  \hfill
  \begin{subfigure}{0.5\linewidth}
    \includegraphics[height=4.0cm]{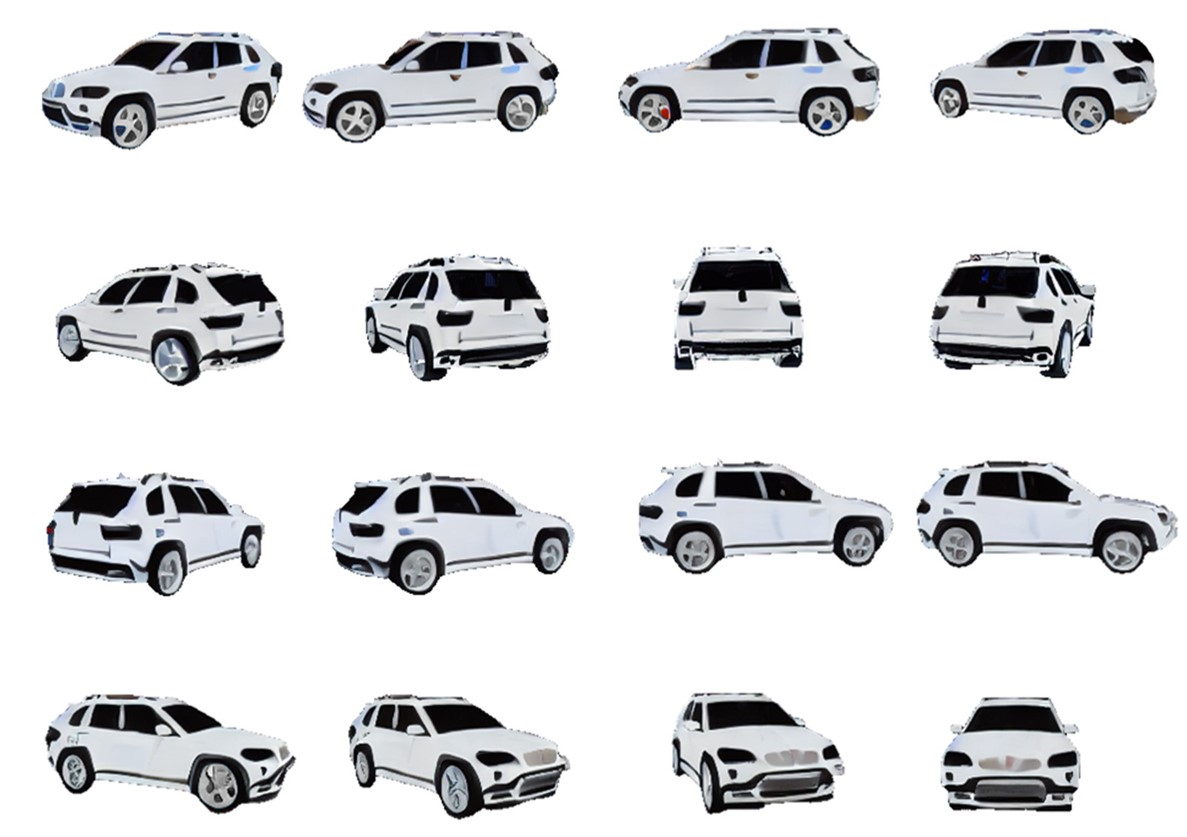}
    \caption{Multi-view structures generated by multi-expert DMs.}
    \label{162}
  \end{subfigure}
  \caption{Comparison of multi-view structures generated by one single DM and multi-expert DMs. The same prompt regarding a BMW X5 is provided for the two methods.}
  \label{16}
\end{figure}

\noindent\textbf{Multi-expert DMs vs One Single DM.}
An alternative to the multi-expert DMs is to create the 16 multi-views with one single text-to-image DM. However, we experimentally found that it results in a less effective learning outcome. A visual comparison between a single text-to-image DM and the multi-expert DMs trained with the same experiment setup is presented in Fig. \ref{16}, where Fig. \ref{161} exhibits inconsistent structural details generated by a single DM. In addition, the overall image quality of a single DM is inferior to that of the multi-expert DMs shown in Fig. \ref{162}. A quantitative ablation study is presented in Sec. \ref{ab}. 

\subsection{Appearance Generation} \label{appear}
\begin{figure}[tb]
  \centering
  \includegraphics[height=4.5cm]{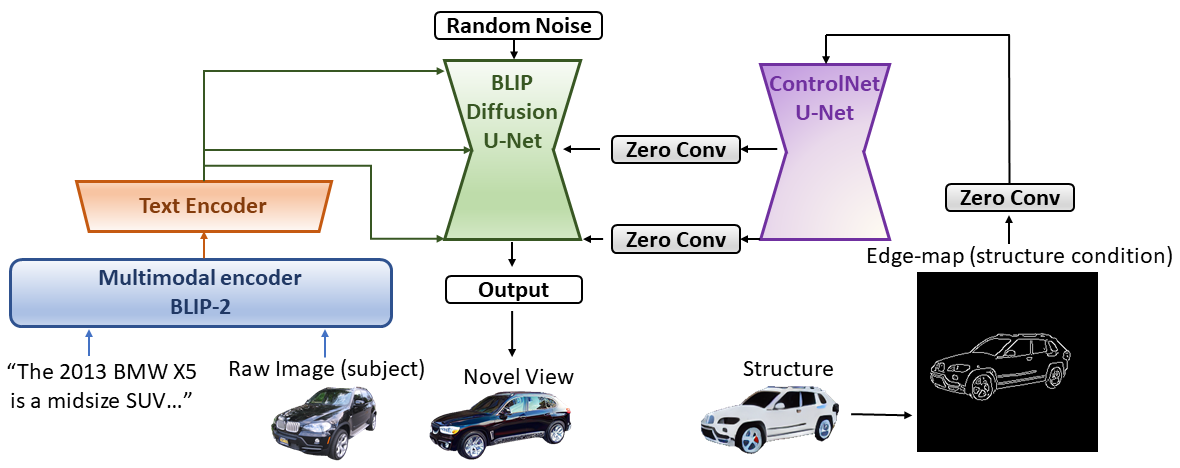}
  \caption{An illustration of the subject-driven structure-controlled generation.
  }
  \label{gen}
\end{figure}




\noindent\textbf{Appearance information extraction.} The structure generation will lead to image appearances in ShapeNet style. To render photorealistic appearances resembling the original vehicle, we need to extract the appearance information. However, this task is challenging due to potential occlusion and tricky viewing angles in the raw images. Therefore, we still apply the VQA model \cite{blip2} to tackle this problem. Yet instead of employing the entire VQA framework, we only utilize its multimodal encoder to encode both the raw image (segmented by SAM \cite{sam} to eliminate the effect of the background) and the prompt. This allows us to robustly extract the appearance information from in-the-wild observations.\par
\noindent\textbf{Novel views rendering.} To utilize the extracted appearance information, we follow previous subject-image-driven generation methods \cite{blipd,dreambooth} and transform the output of the multimodal encoder into text embeddings of a text-to-image DM \cite{Stable}. Thus, the output of the text-to-image DM is controlled by the raw image. Furthermore, considering that only the geometry of the multi-view structures is beneficial, we extract the geometry information through the Canny edge transformation. To control the generation based on the geometry information, we attach an edge-to-image ControlNet \cite{control} to the UNet of the text-to-image DM. In this way, the text-to-image DM \cite{Stable,blipd} generates photorealistic novel views, which are controlled by the VQA result, the raw image, and the vehicle structure prior. Fig. \ref{gen} presents an illustration of the appearance generation.
%
%
%


\section{Experiments} \label{exp}
In this section, we present qualitative and quantitative results on three datasets, Pascal 3D+ \cite{p3d}, Waymo \cite{waymo}, and Objaverse \cite{obj}. We curated 20 vehicles with diverse structures and appearances from each of the datasets.

\noindent\textbf{Implementation Details} 
To train the multi-expert DMs, we create 16 renderings for each instance in the car taxonomy of ShapeNetV2 \cite{shapenet}. Each 3D model is normalized into a cube of $[-0.5,0.5]^{3}$ and rendered in Blender. The virtual cameras are equally spaced around the object with a distance of 1.5 to the object center and an elevation of 5$^\circ$. For the best prompt accuracy, we empirically apply BLIP-2 \cite{blip2} to the first image to generate the prompt. Every fourth image, starting from the first one, is selected as an anchor view. These anchor views are then used with the prompts to train the text-to-image DM. Each anchor view, in conjunction with its three adjacent views, is used to train one image-to-image DM. We adopt the framework of SD v1.5 \cite{Stable} to build up the multi-expert DMs and fine-tune the pretrained text-to-image SD and image-to-image DM \cite{pix2pix} with our datasets. Each DM is trained for 50 epochs with the Adam optimizer \cite{adam}, using a learning rate of 1e-5 and a batch size of one. For the appearance generation, we adopt the pre-trained BLIP-2 multimodal encoder \cite{blip2}, the text encoder and U-Net of BLIP-Diffusion \cite{blipd}, and an edge-to-image ControlNet \cite{control} to construct the subject-driven structure-controlled generation mechanism.


\noindent\textbf{Metrics.} We follow previous work \cite{nfi,dg} to report FID \cite{fid} and the CLIP Similarity \cite{clipsim}. In addition, we report the Image-Text Contrastive (ITC) score proposed in BLIP-2 \cite{blip2} and the VQA score proposed in \cite{score} as metrics to evaluate the matching between the generated views and the VQA results. \par

\noindent\textbf{Competitors.} We compare our method with three state-of-the-art methods, including NFI \cite{nfi}, Zero123XL \cite{zero,obj} and DG \cite{dg}. Particularly, NFI is pretrained on Pasacal 3D+ \cite{p3d}, and Zero123XL and DG are pretrained on Objaverse \cite{obj}. Our method only learned vehicle structures from ShapeNetV2 \cite{shapenet} while inheriting prior knowledge from DMs \cite{Stable,blipd} and the VQA model \cite{blip2}. Note that the superiority of our method is not derived from training on ShapeNetV2 \cite{shapenet} but from the method itself. Please refer to the supplementary material for details.

\begin{figure}[tb]
  \centering
  \includegraphics[height=8.0cm]{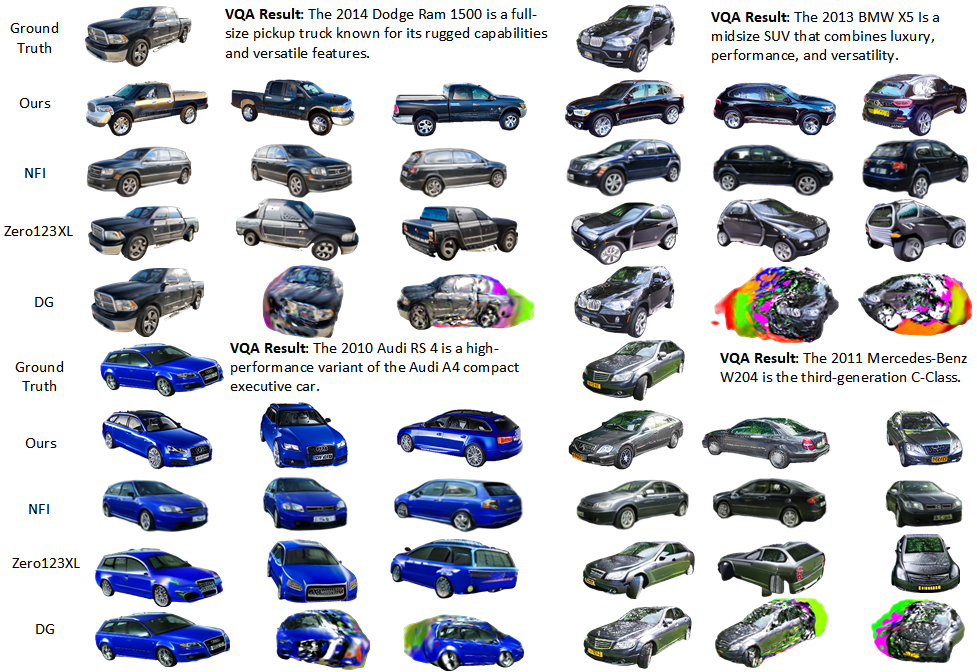}
  \caption{Comparison with state-of-the-art methods (NFI \cite{nfi}, Zero123XL \cite{zero}, and DG \cite{dg}) on the Pascal 3D+ dataset \cite{p3d}.
  }
  \label{p3d}
\end{figure}

\begin{table}[tb]
  \caption{Comparison with the state-of-the-art methods on Pascal 3D+ \cite{p3d}.
  }
  \label{tab:headings}
  \centering \resizebox{0.65\columnwidth}{!} {
  \begin{tabular}{@{}lllll@{}}
    \toprule
    Method  & ITC score $\uparrow$ & \quad  CLIP similarity $\uparrow$  \quad &  FID $\downarrow$ \quad &VQA score $\uparrow$  \\
    \midrule
    Ground Truth \quad \quad & 0.401 \quad \quad & \quad  -- & --  \quad \quad & --\\
    
    DG \cite{dg} \quad \quad & 0.268  \quad \quad & \quad 0.704 \quad \quad & 269.24 \quad \quad & 0.519\\
    Zero123XL \cite{zero} \quad \quad & 0.270  & \quad 0.750 &  193.27 \quad \quad & 0.506 \\
    NFI \cite{nfi} \quad \quad & 0.284 &  \quad 0.784 &  129.43 \quad \quad & 0.599\\
    Ours \quad \quad & 0.380 & \quad 0.856 &  117.49 \quad \quad & 0.903\\
  \bottomrule
  \label{tab1}
  \end{tabular} }
\end{table}

\subsection{Results on Pascal 3D+}
The qualitative and quantitative comparisons of our approach against the state-of-the-art methods are presented in Fig. \ref{p3d} and Table \ref{tab1}.
As seen in Fig. \ref{p3d}, our method visually yields the best quality. Take the first Dodge Ram 1500 pick-up truck as an example, since the observation is obtained from a challenging view, the trunk appears as a small part. As a result, NFI \cite{nfi} ignores the curve between the body and trunk and mistakes it as an SUV. Zero123XL \cite{zero,obj} presents a decent appearance at the original viewing angle, but its geometry estimation fails at other views due to a lack of zero-prediction capability. In particular, it takes the trunk as a small part compared to the body thus the geometry rendered from other views looks unrealistic. DG \cite{dg} shows perfect rendering at the original view as it employs the raw image to train the GS \cite{gs}. However, it utilizes SDS from Zero123XL to reconstruct unseen views, resulting in unsuccessful novel view rendering as Zero123XL fails to provide correct geometry and appearance information. In comparison, our method successfully recognizes the vehicle's model and manufacturer, and thus, it generates the correct geometry of a pick-up truck while rendering photorealistic appearances resembling the raw image. Moreover, as shown in Table \ref{tab1}, the fidelity of our approach outperforms the other state-of-the-art methods quantitatively. Specifically, our method achieves the best ITC score close to the ground truth and the highest VQA score. This indicates that the novel views generated by our approach keep the most semantic information from the perspective of Vision-Language \cite{blip,vqa2,blip2}.

\begin{figure}[tb]
  \centering
  \includegraphics[height=4.5cm]{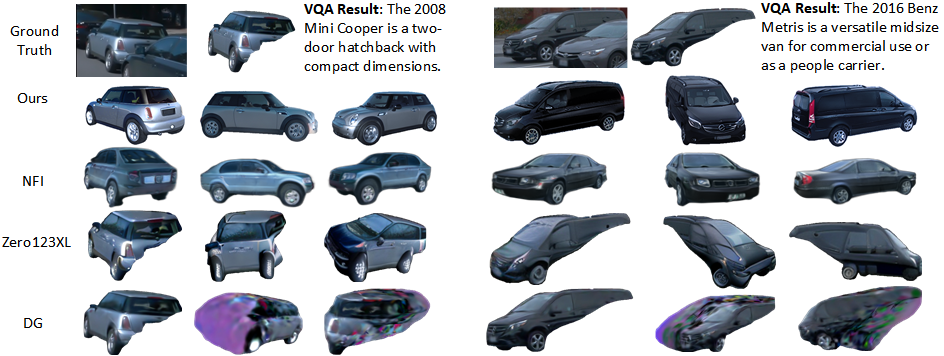}
  \caption{Comparison with state-of-the-art methods (NFI \cite{nfi}, Zero123XL \cite{zero}, and DG \cite{dg}) on the Waymo dataset \cite{waymo}.
  }
  \label{waymo}
\end{figure}

\begin{table}[tb]
  \caption{Comparison with the state-of-the-art methods on Waymo \cite{waymo}.
  }
  \label{tab:headings}
  \centering \resizebox{0.65\columnwidth}{!} {
  \begin{tabular}{@{}lllll@{}}
    \toprule
   Method  & ITC score $\uparrow$ & \quad  CLIP similarity $\uparrow$  \quad &  FID $\downarrow$ \quad \quad &VQA score $\uparrow$   \\
    \midrule
 Ground Truth \quad \quad & 0.422 \quad \quad & \quad  -- & --  \quad \quad & --\\
    DG \cite{dg} &  0.282 \quad \quad & \quad 0.819 & 350.61& 0.644\\
    Zero123XL \cite{zero} & 0.306  & \quad 0.835 & 251.59& 0.589\\
    NFI \cite{nfi} & 0.298  &  \quad 0.829 & 188.23& 0.194 \\
    Ours & 0.418  & \quad  0.840 & 163.40 & 0.854 \\
  \bottomrule
  \label{tab2}
  \end{tabular}}
\end{table}

\subsection{Results on Waymo} 
The qualitative and quantitative results are presented in Fig. \ref{waymo} and Table \ref{tab2} respectively. In this experiment, none of the competitors are pretrained on Waymo \cite{waymo}, ensuring a fair comparison in terms of zero-shot prediction. As seen in Fig. \ref{waymo}, NFI \cite{nfi} is pretrained on Pascal 3D+ \cite{p3d}, where occlusion exists in some instances. Consequently, NFI is able to generate a vehicle, albeit with incorrect geometry. Zero123XL \cite{zero,obj} and DG \cite{dg} cannot handle occlusion at all. They interpret occluded parts as non-existing and even fail to generate car-like objects. In contrast, our method can generate complete and accurate structures along with high-fidelity appearances, even from occluded observations. As shown in Table \ref{tab2}, our method yields the best performance. 

\begin{figure}[tb]
  \centering
  \includegraphics[height=4.5cm]{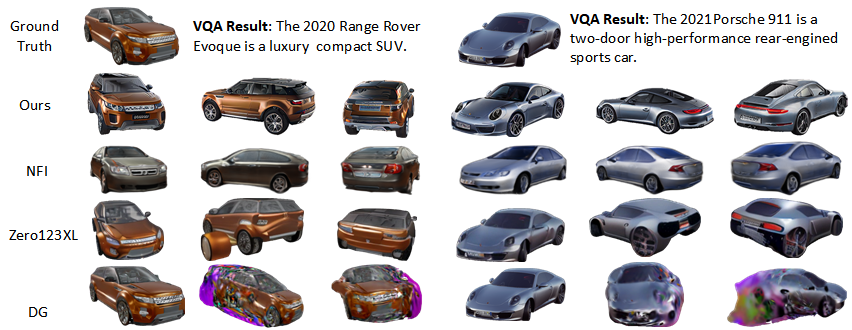}
  \caption{Comparison with state-of-the-art methods (NFI \cite{nfi}, Zero123XL \cite{zero}, and DG \cite{dg}) on the Objaverse dataset \cite{obj}.
  }
  \label{obj}
\end{figure}

\begin{table}[t]
  \caption{Comparison with the state-of-the-art methods on Objaverse \cite{obj}.
  }
  \label{tab:headings}
  \centering \resizebox{0.65\columnwidth}{!} {
  \begin{tabular}{@{}lllll@{}}
    \toprule
    Method  & ITC score $\uparrow$ & \quad  CLIP similarity $\uparrow$  \quad &  FID $\downarrow$ \quad & \quad VQA score $\uparrow$  \\
    \midrule
 Ground Truth \quad \quad & 0.429 \quad \quad & \quad  -- & --  \quad \quad & \quad --\\  
    DG \cite{dg} & 0.269 & \quad 0.761 &  296.39 & \quad 0.516\\
    Zero123XL \cite{zero} & 0.319  & \quad 0.812 & 175.18 & \quad 0.366 \\
    NFI \cite{nfi} & 0.289  &  \quad 0.772 & 146.06 & \quad 0.669\\
    Ours & 0.412  & \quad 0.872 & 114.75 & \quad 0.838\\
  \bottomrule
  \label{tab3}
  \end{tabular}}
\end{table}

\subsection{Results on Objaverse} 
The qualitative and quantitative results are presented in Fig. \ref{obj} and Table \ref{tab3} respectively. Zero123XL \cite{zero,obj} and DG \cite{dg} have a certain advantage in this test as they are pretrained on Objaverse \cite{obj}. For NFI \cite{nfi} and our method, it is still a test for zero-shot prediction. As depicted in Fig. \ref{obj}, the inputs consist of observations from relatively challenging viewing angles compared to a simple side view. Take the Range Rover Evoque as an example, as the color dark orange is not common in Pascal 3D+ \cite{p3d}, NFI fails to create the correct appearance while the structure is complete and close to the raw image. The novel views generated by Zero123XL, while exhibiting a similar appearance to the raw image, suffer from severe structural defects, such as an additional wheel. Since DG \cite{dg} relies on Zero123XL to reconstruct unseen views, the novel views given by DG are messy. In comparison, our method creates novel views with precise structures and photorealistic appearances akin to the raw image. Consequently, our method outperforms the other state-of-the-art methods as presented in Table \ref{tab3}.

\begin{table}[tb]
  \caption{Ablation study of the multi-expert DMs.
  }
  \label{tab:headings}
  \centering \resizebox{0.6\columnwidth}{!} {
  \begin{tabular}{@{}llll@{}}
    \toprule
    Method &  ITC score $\uparrow$  \quad \quad &   CLIP similarity $\uparrow$ \quad \quad  & FID $\downarrow$ \\
    \midrule
Single DM (50 epochs) & 0.272 &  0.781 & 192.55\\
   Single DM (100 epochs) & 0.283 &  0.806 & 189.04\\
    Multi-expert DMs (50 epochs) & 0.333 & 0.835 & 122.91\\
  \bottomrule
  \end{tabular} \label{tab4}}
\end{table}
\begin{figure}[tb]
  \centering
  \includegraphics[height=7.2cm]{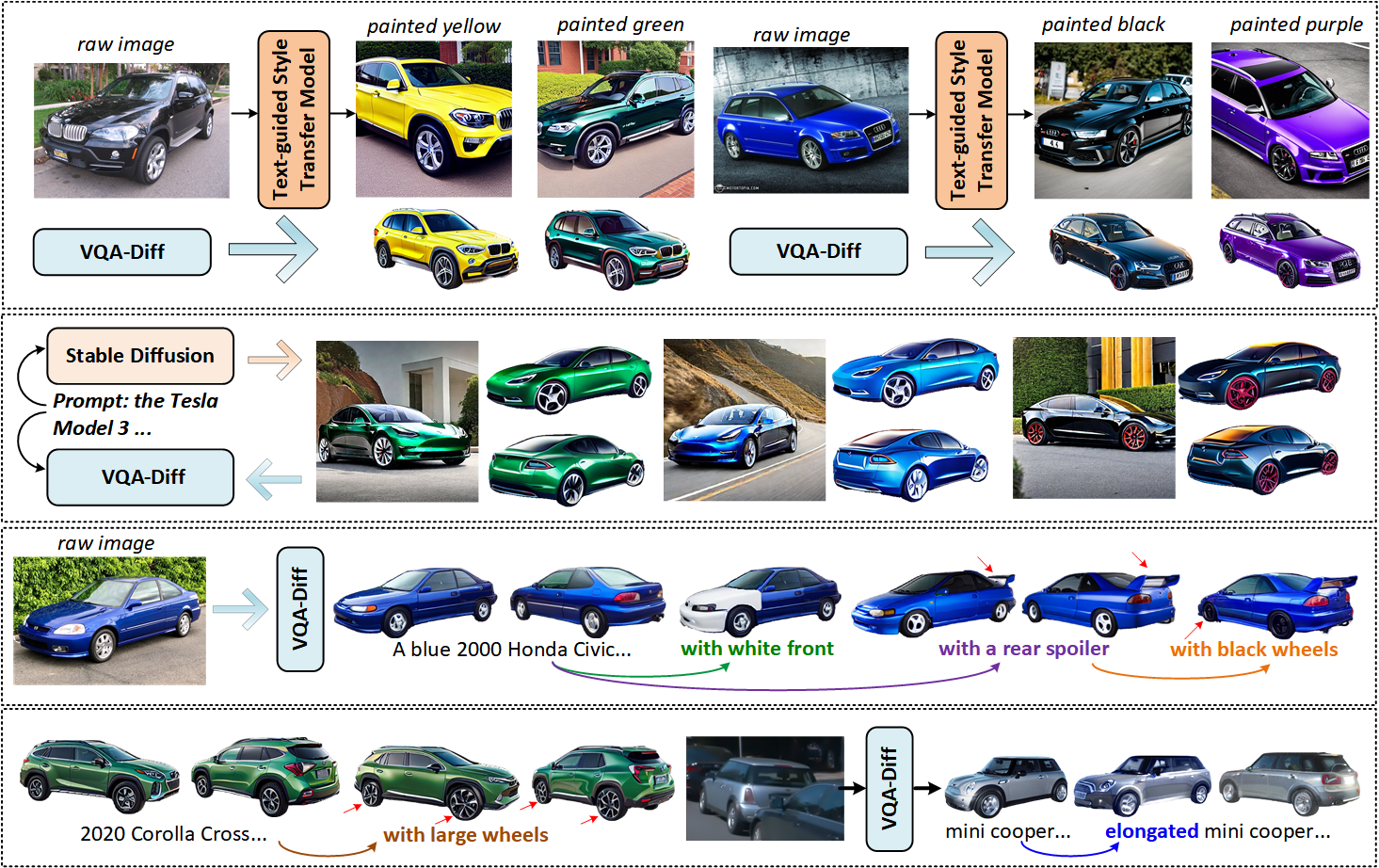}
  \caption{Evaluation on generating extended diverse 3D vehicle assets through collaboration with the text-guided-style-transfer model \cite{pix2pix,blipd}, text-to-image generation model \cite{Stable}, and text and shape guided object inpainting model \cite{brush}.
  }
  \label{diverse}
\end{figure}

\subsection{Ablation Study} \label{ab}
We conducted an experiment to compare the effectiveness of multi-expert DMs with a single text-to-image DM, from which all 16 views are generated. We take the ShapeNet renderings of a BMW X5 as the ground truth and provide the corresponding prompt to our multi-expert DMs and the single DM. Table \ref{tab4} shows the quantitative comparison. As seen, by adopting multi-expert, the model can learn structure generation more efficiently while achieving a better performance. The analysis of this experiment and ablation study on the structure training dataset and ControlNet are provided in the supplementary material.

\subsection{Extended Diverse 3D Vehicle Assets Generation }
In addition to creating 3D vehicle assets from in-the-wild observations, we introduce complementary solutions, as depicted in Fig. \ref{diverse}, for generating extended diverse vehicle assets to build a comprehensive asset bank. In particular, by collaborating with text-guided style transfer models \cite{pix2pix,blipd}, text-to-image generative models \cite{Stable}, and object inpainting models \cite{brush}, we can alter the colors of vehicles, generate vehicles merely through designed prompts, and create intricate vehicles by providing fine-grained prompts (such as spoiler, wheel, shape and part). More details and results are presented in the supplementary material.

\subsection{Limitation}
Despite the promising results, our method still has limitation. Our method is elaborately designed for vehicle asset generation, so when extending it to more generic objects, it may not always perform well (please refer to the supplementary material for the success and failure cases). We conjecture that the main reason is that vehicles are a specific type of object whose structures can be determined by specifying the model, make, and production year, which allows VQA model to provide geometry constraints by generating precise prompts regarding key information. However, it is challenging for the VQA model to constrain the structure of a generic object (such as a teddy bear). With the ongoing development of VQA, extending the proposed method to more objects shows promise.



\section{Conclusion}
In this work, we propose a new VQA-Diff framework to facilitate novel view rendering from in-the-wild vehicles for autonomous driving. The core idea is to integrate the robust zero-shot prediction ability of the VQA model into novel view rendering. Thus, VQA-Diff can deal with in-the-wild observations with occlusions and a challenging viewing angle. 
We first design the VQA processing to extract information about the vehicle from a deeper understanding than the previous image-to-3D methods. Then, we develop the multi-expert DMs for learning to generate vehicle structures based on the VQA result from a synthesis dataset. Finally, we apply the subject-driven structure-controlled generation mechanism to transform the multi-view structures into photorealistic novel view renderings. Qualitative and quantitative evaluations on Pascal 3D+, Waymo, and Objaverse
show the superiority of our approach over the state-of-the-art methods.

\clearpage  

%
%


\section* {Supplementary Material}

\subsection*{A. Overview}
This material includes both supplementary quantitative and qualitative experimental results, along with additional information such as implementation details, and discussions to complement the main paper.

\subsection*{B. Extended 3D Vehicle Assets Generation}
In this section, we introduce solutions to generate extended 3D vehicle assets with the proposed method.\par
\begin{figure}[t]
  \centering
  \includegraphics[height=11.0cm]{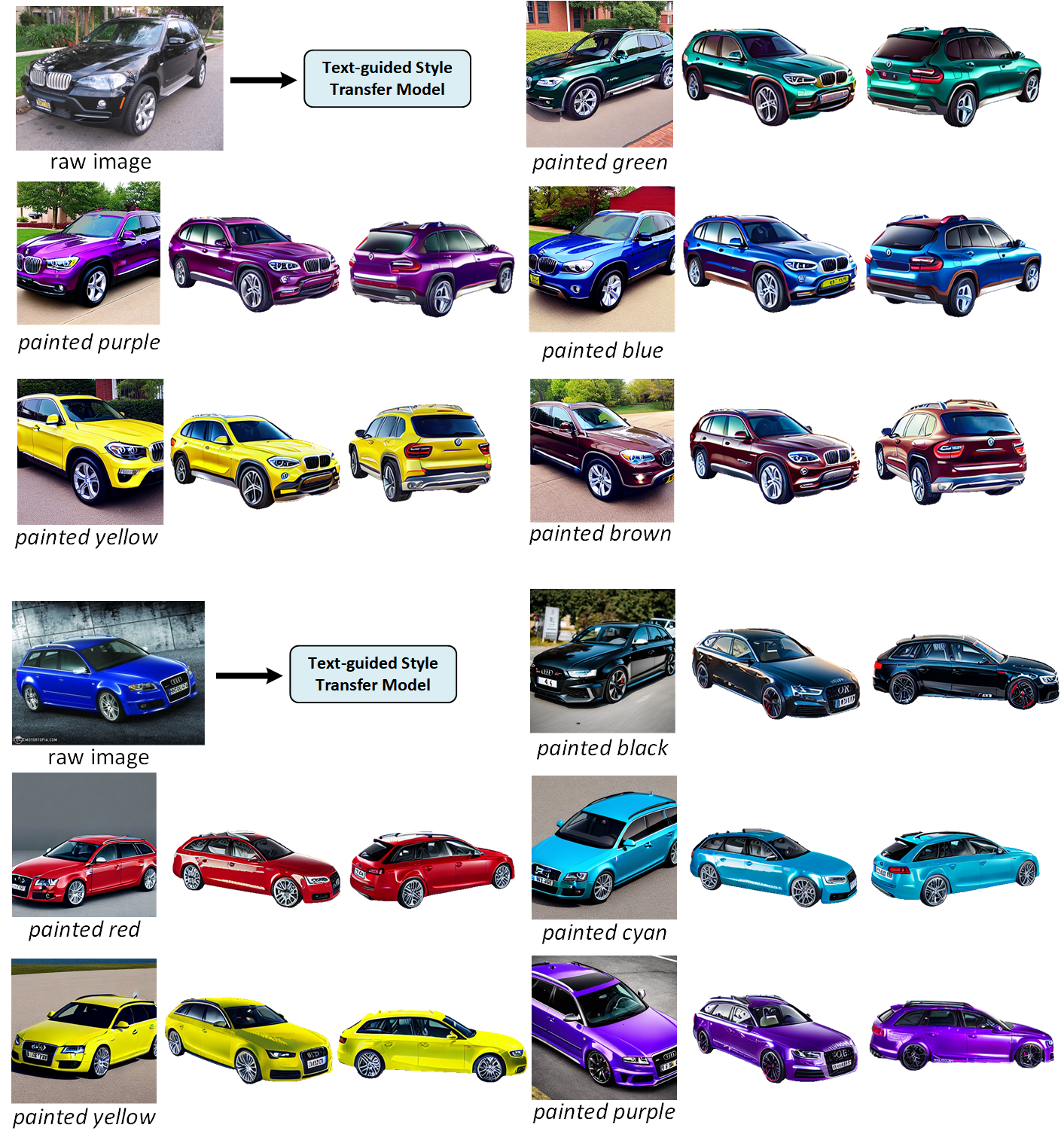}
  \caption{We propose applying text-guided style transfer models \cite{pix2pix,blipd} to alter the vehicle's paint in the raw image, utilizing prompts like "painted brown". Then, we use the proposed method to generate photorealistic renderings of the same vehicle model with different paintings.
  }
  \label{paint}
\end{figure}

\noindent\textbf{Cooperating with Text-guided Style Transfer Models.} While the proposed method effectively addresses the challenge of rendering novel views from in-the-wild observations, it is also desired to generate diverse appearances to construct a comprehensive 3D asset bank. Thus, we propose to use the text-guided style transfer models \cite{pix2pix,blipd} to transform the paint of the raw vehicle, and then generate extended 3D vehicle assets through the proposed method. An illustration of this solution is presented in Fig. \ref{paint}.

\begin{figure}[t]
  \centering
  \includegraphics[height=9.0cm]{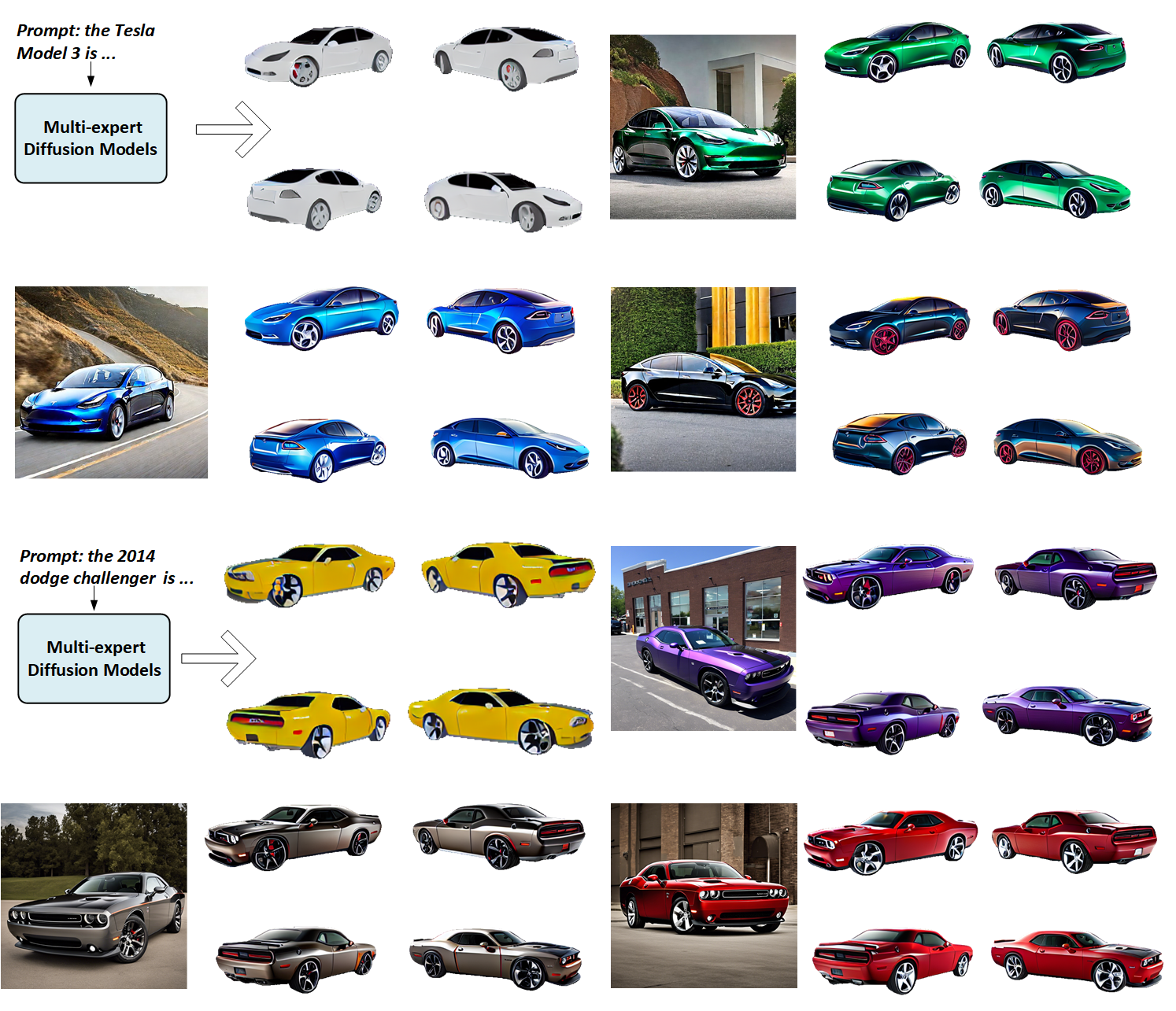}
  \caption{We propose applying text-to-image models, such as Stable Diffusion \cite{Stable}, to generate reference images for our method. Consequently, our approach can produce photorealistic renderings of vehicles without relying on in-the-wild observations. In this example, the reference images are generated by Stable Diffusion \cite{Stable} from the manually designed prompts. Note that the multi-expert DMs did not learn the structures of a Tesla Model 3 as ShapeNetV2 \cite{shapenet} does not encompass this model.
  }
  \label{sub1}
\end{figure}

\begin{figure}[t]
  \centering
  \includegraphics[height=5.0cm]{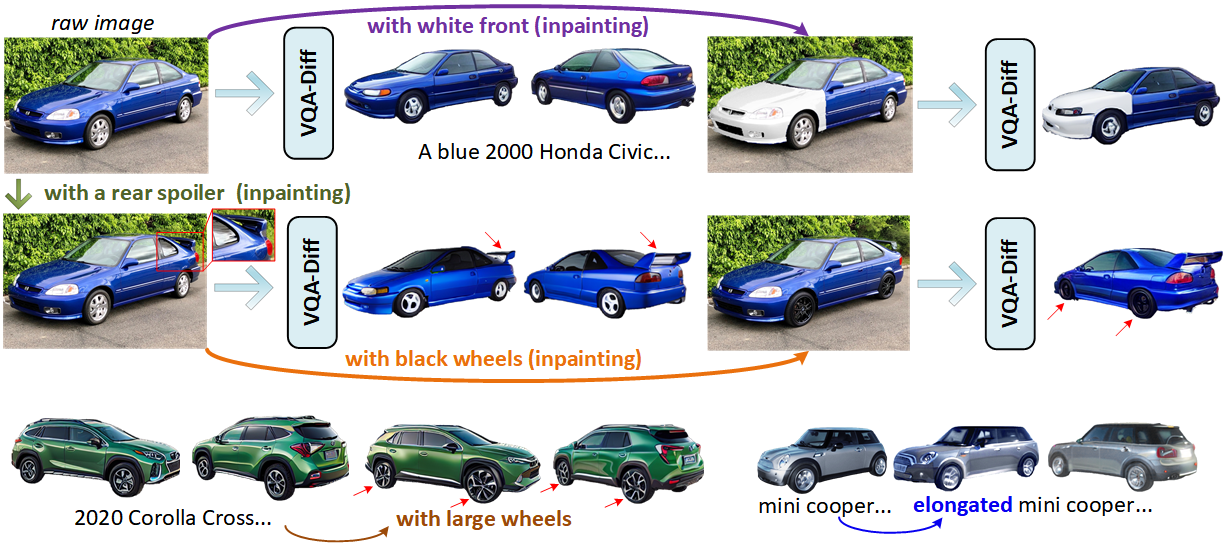}
  \caption{ We propose applying text and shape guided object inpainting models, such as SmartBrush \cite{brush}, to generate intricate vehicles with desired details. Utilizing fine-grained prompts, the inpainting model creates a new reference image with desired details. Then, VQA-Diff generates intricate vehicle assets based on the new reference image and fine-grained prompt.
  }
  \label{fine}
\end{figure}
\begin{figure}[h]
  \centering
  \includegraphics[height=6.5cm]{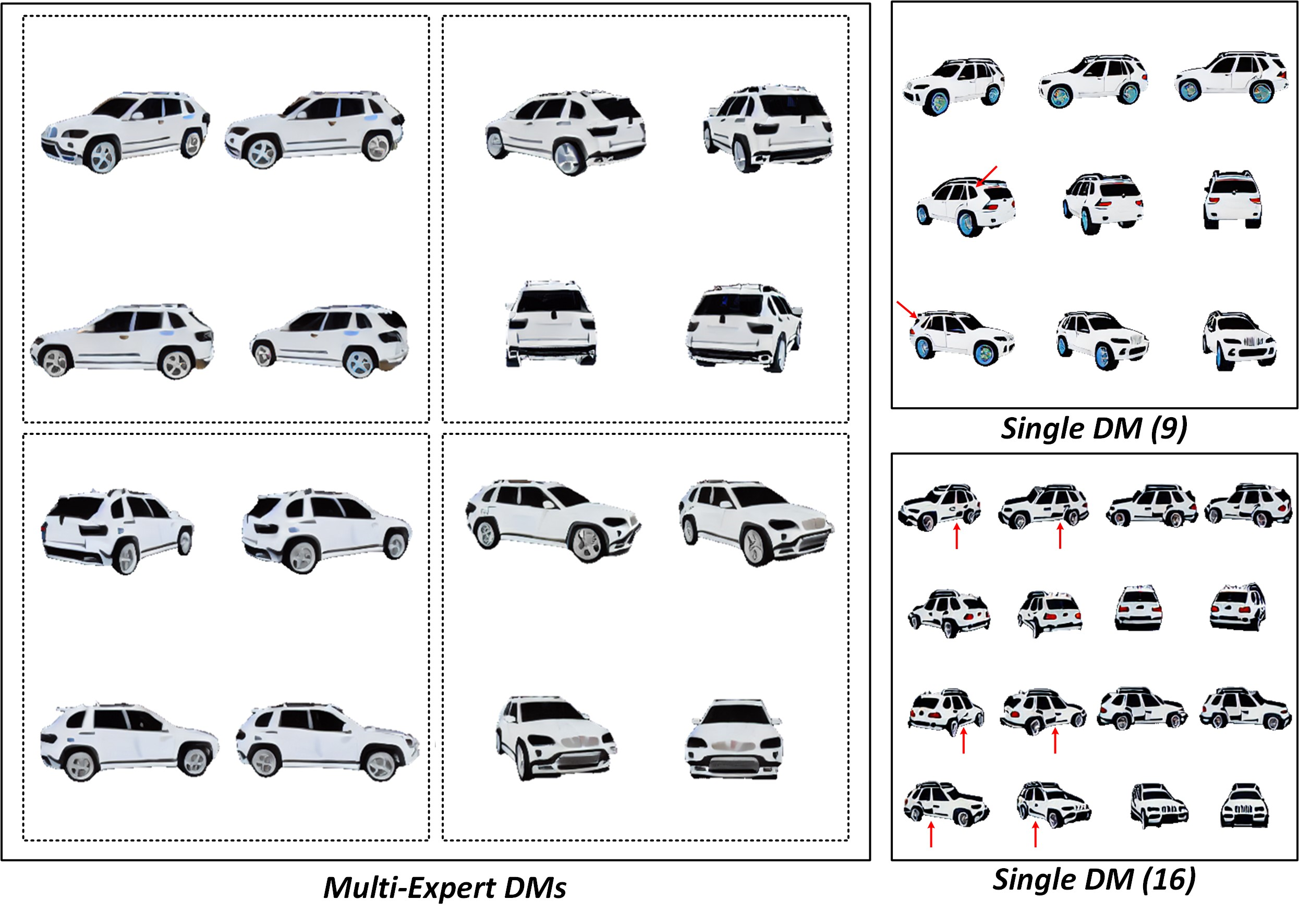}
  \caption{A visual comparison of the generated multi-view structures using different DM schemes.
  }
  \label{views}
\end{figure}
\begin{figure}[h]
  \centering
  \includegraphics[height=6.5cm]{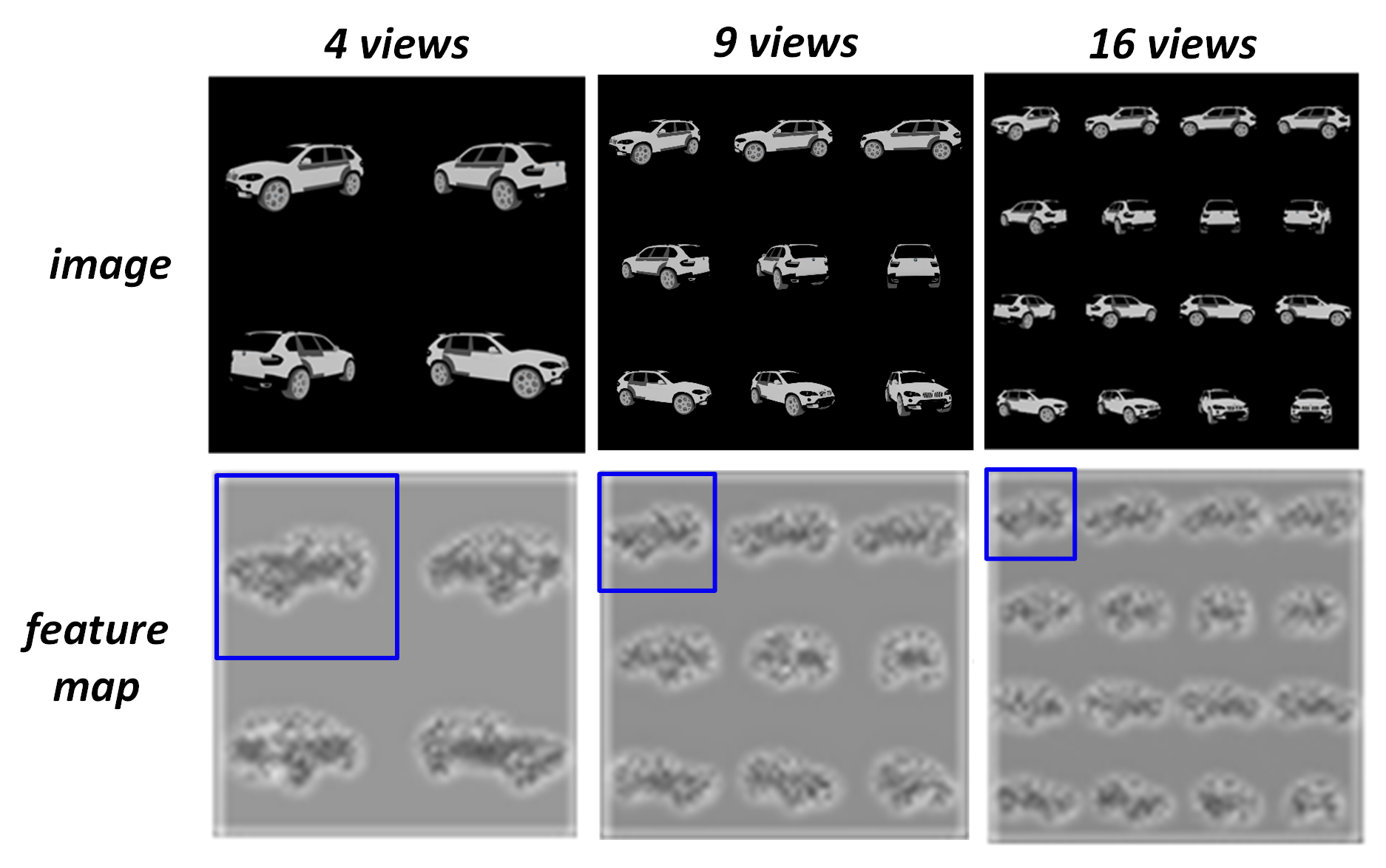}
  \caption{An illustration of how the number of views affects the feature map. 
The representations of an individual view are denoted by blue boxes.
  }
  \label{feature}
\end{figure}
\begin{figure}[h]
  \centering
  \includegraphics[height=2.5cm]{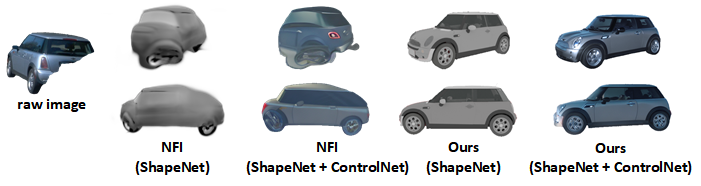}
  \caption{A visual comparison of NFI \cite{nfi} and our method trained on ShapeNetV2 \cite{shapenet} w/ and w/o ControlNet\cite{blipd,control}.
  }
  \label{shapenet}
\end{figure}
\begin{figure}[h]
  \centering
  \includegraphics[height=2.5cm]{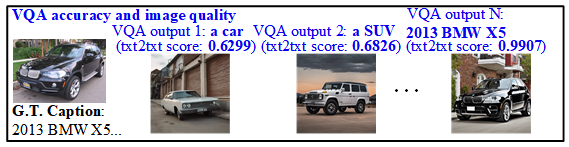}
  \caption{Evaluation of the generated images during our question design process.}
  \label{vqaa}
\end{figure}
\begin{figure}[h]
  \centering
  \includegraphics[height=2.5cm]{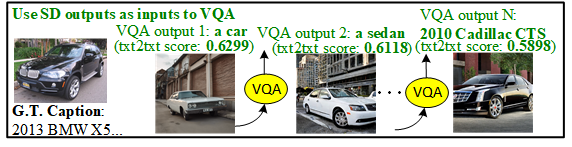}
  \caption{An illustration depicting the effect of an alternative question design solution.
  }
  \label{vqab}
\end{figure} 
\begin{figure}[h]
  \centering
  \includegraphics[height=1.5cm]{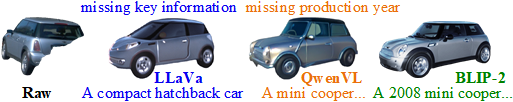}
  \caption{Comparison of adopting different VQA models \cite{llava,qwen,blip2} in VQA-Diff.
  }
  \label{r2}
\end{figure} 
\begin{figure}[h]
  \centering
  \includegraphics[height=7cm]{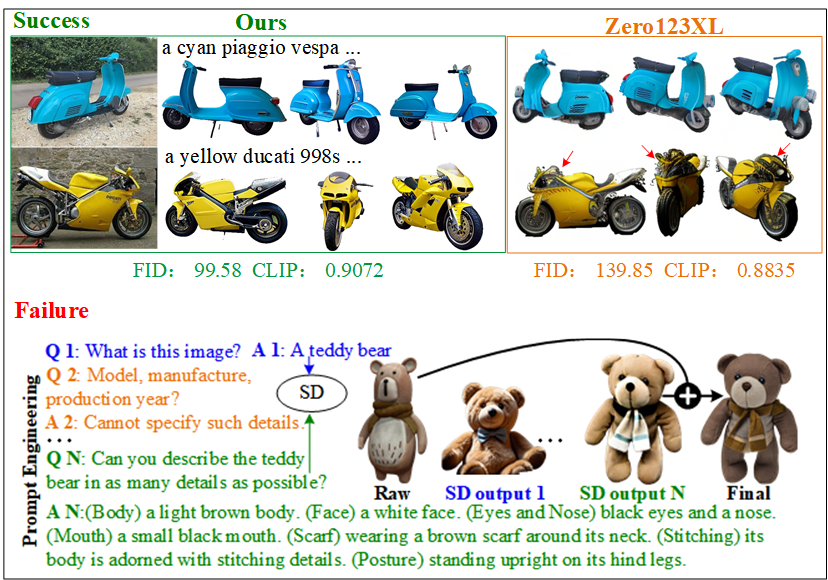}
  \caption{Success and failure cases of other categories.
  }
  \label{lim}
\end{figure} 

\noindent\textbf{Cooperating with Text-to-image Models.} We cannot guarantee that the data collected on the road covers all the wanted car models. Hence, to build a complete 3D vehicle assets bank, we also want to generate vehicle assets without requiring in-the-wild observations. Therefore, we propose to use text-to-image generative models, such as Stable Diffusion (SD) \cite{Stable}, to produce reference images for VQA-Diff directly. In this way, the proposed method can generate 3D vehicle assets based on prompts. An illustration of this solution is depicted in Fig. \ref{sub1}. 
\par
\noindent\textbf{Cooperating with Text and Shape Guided Object Inpainting Models.} VQA-Diff can cooperate with text and shape-guided object inpainting models \cite{brush} to generate intricate vehicles by providing fine-grained prompts, such as spoiler, wheel, shape, and part. The results are presented in Fig. \ref{fine}. In particular, we adopt the inpainting model to change the details of the raw vehicle by providing a fine-grained prompt, such as adding a rear spoiler, and specifying the desired inpainting region. Simultaneously, we modify the VQA result with fine-grained words (\eg, "with a rear spoiler") to guide the generation of the desired geometry with VQA-Diff. Finally, the newly obtained reference image and the multi-views are utilized to create intricate vehicles with desired details. 
\par
\noindent\textbf{Remark.} Note that the reference images shown in Fig. \ref{paint} and Fig. \ref{sub1}, while photorealistic, are not 3D vehicle assets, as their object poses are not controllable. In contrast, the renderings created by our method have controlled object poses. In addition, we want to emphasize that generating vehicle assets from real-world observations holds particular significance in constructing a simulation environment for autonomous driving \cite{sim1,sim2}.  Because the appearances of vehicles generated in this manner closely resemble those found in data collected on the road. 

\subsection*{C. Effect of Number of DMs}
In this section, we explore the effect of adopting different DM schemes as a supplementary experiment to the main paper. Specifically, we compare the proposed multi-expert DMs with a single DM. In the single DM scheme, the model learns to generate all desired multi-view structures in one image. "Single DM (16)" and "Single DM (9)" imply that there are either 16 views or 9 views within a single image. All the models are obtained through fine-tuning a pretrained Stable Diffusion v1.5 \cite{Stable} on the ShapeNetV2 \cite{shapenet} for 50 epochs with a learning rate of 1e-5 and a batch size of one. We input the same prompt regarding a BMW X5 into all the models and compare the generated images with the multi-view images of the BMW X5 in the training data. The qualitative and quantitative comparisons are presented in Fig. \ref{views} and Table \ref{subtab}. As seen in Table \ref{subtab}, the multi-expert DMs yield better performance than the single DM schemes. Moreover, as pointed out by the red arrows in Fig. \ref{views}, there are inconsistent parts in the views generated by the single DM schemes.
We provide our analysis in the following. The SD Model \cite{Stable} is a Latent Diffusion Model (LDM). The learning of the LDM is supervised by both text embeddings (derived from the prompts) and feature maps (extracted from the raw images). Given that the prompts provided to the models are the same, we analyze the $4 \times 64 \times 64$ feature map encoded by VAE \cite{clip}. Considering that the resolution of the feature map is fixed in VAE, as seen in Fig. \ref{feature}, the representation quality of a single view degrades as the number of views increases. This indicates that the single DM has to learn from a supervision with a worse quality and the performance will be inferior to our method. 
\begin{table}[tb]
  \caption{Ablation study of the multi-expert DMs.
  }
  \label{tab:headings}
  \centering
  \centering \resizebox{0.6\columnwidth}{!} {\begin{tabular}{@{}llll@{}}
    \toprule
    Method &  ITC score $\uparrow$  \quad \quad &   CLIP similarity $\uparrow$ \quad \quad  & FID $\downarrow$ \\
    \midrule
Single DM (16) & 0.272 &  0.781 & 192.55\\
   Single DM (9) & 0.311 &  0.811 & 150.31\\
    Multi-expert DMs  & 0.333 & 0.835 & 122.91\\
  \bottomrule
  \end{tabular} \label{subtab}}
\end{table}

\begin{table}[tb]
  \caption{Ablation study of the structure training dataset and ControlNet.
  }
  \label{tab:headings}
  \centering
  \centering \resizebox{0.9\columnwidth}{!} {\begin{tabular}{@{}llll@{}}
    \toprule
    Method &  ITC score $\uparrow$  \quad \quad &   CLIP similarity $\uparrow$ \quad \quad  & FID $\downarrow$ \\
    \midrule
NFI \cite{nfi}  (ShapeNet \cite{shapenet}) & 0.210 &  0.744 & 428.35\\
NFI \cite{nfi}  (ShapeNet \cite{shapenet} + ControlNet \cite{blipd,control}) & 0.261 &  0.761 & 267.47\\
    Ours (ShapeNet \cite{shapenet})  & 0.403 & 0.831 & 186.98\\
    Ours (ShapeNet \cite{shapenet} + ControlNet \cite{blipd,control})  & 0.418 & 0.840 & 163.40\\
  \bottomrule
  \end{tabular} \label{subtab2}}
\end{table}
\subsection*{D. Impact of Structure Training Dataset and ControlNet}
In this section, we study the impact of the structure training dataset and ControlNet. We compare all the methods on Waymo \cite{waymo}. First, to demonstrate that the superiority of VQA-Diff is not brought by training on ShapeNetV2 \cite{shapenet} but the method itself, we compare the NFI \cite{nfi} trained on ShapeNetV2, which is denoted by NFI (ShapeNet), with our method without adopting ControlNet, which is denoted by Ours (ShapeNet). As presented in Table \ref{subtab2}, although both models are trained on ShapeNetV2, our method outperforms NFI as our method generates the correct geometry using the VQA result. The appearance of novel views in our method is created by ControlNet \cite{blipd,control}. Thus, for a fair and comprehensive comparison, we also apply the ControlNet to the result of NFI (ShapeNet), which is represented by NFI (ShapeNet+ControlNet). As seen in Table \ref{subtab2}, although the performance is boosted by applying the ControlNet, NFI (ShapeNet+ControlNet) is still inferior to Ours (ShapeNet) because the geometry is wrong. Fig. \ref{shapenet} shows the visual comparison.

\subsection*{E. Impact of Question Design and VQA Model }
Fig. \ref{vqaa} shows the evaluation of the generated images during the question design process. In particular, we employ the CLIP txt2txt score \cite{clip,make} to evaluate the matching between the VQA prediction and the ground truth caption. In particular, the original image is used to update the prompt, and we compute the txt2txt score each time we obtain an answer from the VQA model. As seen, the score increases and the image quality improves until reaching convergence, where the txt2txt score is high and the generated image refers to the same vehicle as the raw image. \par
Compared to the question design solution proposed in VQA-Diff, one might consider an alternative method: iteratively using SD outputs as inputs to VQA, where the SD outputs are used to update the prompt. We present the results of this solution in Fig. \ref{vqab}. As seen, from the second step, SD predicts wrong images as inputs to VQA, so using SD outputs as inputs to VQA yields worse score and image quality. \par
The effect of adopting different VQA models in VQA-Diff is illustrated in Fig. \ref{r2}. Although there are multiple SOTA VQA models, such as LLaVA \cite{llava}, Qwen VL \cite{qwen}, and BLIP-2 \cite{blip2}, we chose BLIP-2 because we found it capable of recognizing more vehicle cues. As seen in Fig. \ref{r2}, LLaVa fails to provide key information while QWen VL fails to provide production year, resulting in inferior renderings, although they might outperform BLIP-2 in other VQA tasks.
We conjecture this is attributed to pretraining data of VQA models.

\subsection*{F. Application to More Generic Objects}
As mentioned in the main paper, extending the proposed method to generic objects is challenging. We present success and failure cases of other categories in Fig. \ref{lim}. For the success case, as shown in Fig. \ref{lim}, VQA-Diff can be extended to generate some motorbike assets and shows better results compared with those generated by Zero123XL \cite{zero,obj}. However, we also found that VQA predictions are less robust for motorbikes compared to cars, which we conjecture is due to the pretraining data of VQA. Moving on to the failure case, as shown in Fig. \ref{lim}, our approach does not show good result for the teddy bear generation. In particular, we tried to follow the same prompt engineering process for dealing with vehicles to obtain geometry constraints via VQA. Unfortunately, VQA cannot provide key information about the teddy bear (such as model, manufacturer, etc.), so it cannot constrain fine-grained structures. Consequently, although the final description of the teddy bear includes many details and the color of the final rendering is close to the raw image, the structure of the final rendering is still inconsistent with the raw image.

\clearpage

\bibliographystyle{splncs04}
\bibliography{egbib}

\end{document}